\documentclass[journal=jacsat,manuscript=article]{achemso}
\setkeys{acs}{usetitle = true}
\setkeys{acs}{etalmode=truncate, maxauthors=5}
\usepackage[version=3]{mhchem} 

\author{Seongok Ryu}
\affiliation[KAIST]{Department of Chemistry, KAIST, 291 Daehak-ro, Yuseong-gu, Daejeon 34141, Republic of Korea}
\author{Yongchan Kwon}
\affiliation[SNU]{Department of Statistics, Seoul National University, Seoul 08826, Republic of Korea}
\author{Woo Youn Kim}
\affiliation[KAIST]{Department of Chemistry, KAIST, 291 Daehak-ro, Yuseong-gu, Daejeon 34141, Republic of Korea}
\alsoaffiliation[KI4AI]{KI for Artificial Intelligence, KAIST, 291 Daehak-ro, Yuseong-gu, Daejeon 34141, Republic of Korea}
\email{wooyoun@kaist.ac.kr}

\title[An \textsf{achemso} demo]{Uncertainty quantification of molecular property prediction with Bayesian neural networks}
\abbreviations{Deep learning, Bayesian neural network, Uncertainty quantification, AI-safety}

\usepackage{url}
\usepackage{dsfont}
\usepackage{kotex}
\usepackage{color}
\begin{document}

\begin{tocentry}

Some journals require a graphical entry for the Table of Contents.
This should be laid out ``print ready'' so that the sizing of the
text is correct.

Inside the \texttt{tocentry} environment, the font used is Helvetica
8\,pt, as required by \emph{Journal of the American Chemical
Society}.

The surrounding frame is 9\,cm by 3.5\,cm, which is the maximum
permitted for  \emph{Journal of the American Chemical Society}
graphical table of content entries. The box will not resize if the
content is too big: instead it will overflow the edge of the box.

This box and the associated title will always be printed on a
separate page at the end of the document.

\end{tocentry}

\begin{abstract}

Deep neural networks have outperformed existing machine learning models in various molecular applications. 
In practical applications, it is still difficult to make confident decisions because of the uncertainty in predictions arisen from insufficient quality and quantity of training data. 
Here, we show that Bayesian neural networks are useful to quantify the uncertainty of molecular property prediction with three numerical experiments. 
In particular, it enables us to decompose the predictive variance into the model- and data-driven uncertainties, which helps to elucidate the source of errors. 
In the logP predictions, we show that data noise affected the data-driven uncertainties more significantly than the model-driven ones. 
Based on this analysis, we were able to find unexpected errors in the Harvard Clean Energy Project dataset. 
Lastly, we show that the confidence of prediction is closely related to the predictive uncertainty by performing on bio-activity and toxicity classification problems.
\end{abstract}

\section{Introduction}

Modern deep neural network (DNN) models have been used in various molecular applications, such as high-throughput screening for drug discovery \cite{gomes2017atomic, jimenez2018k, mayr2016deeptox, ozturk2018deepdta}, \textit{de novo} molecular design \cite{de2018molgan, gomez2018automatic, guimaraes2017objective, jin2018junction, kusner2017grammar, li2018learning, segler2017generating, you2018graph} and planning chemical reaction \cite{segler2018planning, wei2016neural, zhou2017optimizing}. 
DNNs show comparable or sometimes better performance than traditional approaches grounded on quantum chemical theories in predicting some molecular properties \cite{faber2017prediction, gilmer2017neural, schutt2017schnet, schutt2017quantum, smith2017ani}, if a vast amount of well-qualified data is secured. 
Despite the remarkable potential of DNN models, the direct use of their outputs is sometimes limited because most data in practical applications is likely to involve undesirable problems caused by the lack of both data quality and quantity.

Such data discourages a reliable statistical analysis based on DNN models, since their accuracy critically depends on training data.
For example, \citeauthor{feinberg2018spatial} mentioned that more qualified data should be provided to improve the prediction accuracy on drug-target interactions, which is a key step for drug discovery \cite{feinberg2018spatial}.
The number of ligand-protein complex samples in the PDB-bind database \cite{liu2017forging} is only about 15,000, limiting the development of reliable DNN models.
In order to prepare more qualified data, expensive and time-consuming experiments are inevitable. 
Synthetic data from computations can be used as an alternative, like the Harvard Clean Energy Project set \cite{hachmann2011harvard}, but it often suffers from unintentional errors caused by approximation methods employed.
In addition, data-inherent bias and noise hurt the quality of data.
Tox21 \cite{mayr2016deeptox} and DUD-E dataset \cite{mysinger2012directory} are such examples.
The number of data in the Tox21 dataset is less than 10,000.
There are far more negative samples than positive samples. 
Of various toxic types, the lowest percentage of positive samples is 2.9\% and the highest is 15.5\%. 
For the DUD-E dataset, it is highly imbalanced that the number of decoy samples are almost 50 times larger than that of active samples. 
All of those situations would interrupt developing reliable models. 

It has been stressed in deep learning researches that uncertainty analysis is necessary to address namely the AI-safety problems. \cite{gal2016uncertainty, begoli2019need, mcallister2017concrete}
That is because even though DNNs push the bounds of data-driven approaches, they often make catastrophic decisions. 
The uncertainty analysis has been performed to analyze the processes of decision making with deep neural networks.
\citeauthor{kendall2017uncertainties} studied quantitative uncertainty analysis on computer vision problems by using Bayesian neural networks (BNNs).\cite{kendall2017uncertainties}
They separated model- and data-driven uncertainties, which helps to identify the sources of prediction errors.
It is possible because Bayesian inference allows uncertainty assessments, giving probabilistic interpretations of model outputs. 

In this paper, we propose to exploit BNNs to quantify uncertainties implied in molecular property predictions. 
Previous studies on uncertainty quantification have regarded a predictive variance as a predictive uncertainty \cite{kendall2017uncertainties, kwon2018uncertainty}. 
The predictive uncertainty can be decomposed into (i) an aleatoric uncertainty arisen from data noise and (ii) an epistemic uncertainty arisen from the incompleteness of model \cite{der2009aleatory}. 
We adopt the same method in this study. 
As a DNN model for molecular applications, we use augmented graph convolutional networks (GCNs) \cite{duvenaud2015convolutional, kipf2016semi, ryu2018deeply}. 
In what follows, we briefly introduce BNNs, the uncertainty quantification methods based on Bayesian inference, and the augmented-GCN used in this work. 
Then, we show the results of uncertainty analysis on three experimental studies.
The main results are summarized as follows. 
\begin{itemize}
    \item We first applied the Bayesian GCN to a simple example, the logP prediction of molecules in the ZINC set\cite{irwin2005zinc}, in order to demonstrate the uncertainty quantification in molecular applications.
    As expected, the aleatoric uncertainty increases as the data noise increases, while the epistemic uncertainty slightly depends on the quality of data.  
    \item Second, we evaluate the quality of synthetic data and find erroneous samples fabricated by poor approximations. 
    The Harvard Clean Energy Project (CEP) set \cite{hachmann2011harvard} contains synthetic power conversion efficiency (PCE) values of molecules.
    We noted that molecules with exactly zero values have a conspicuously large aleatoric uncertainty, which have been verified as incorrect annotations.
    \item In the last example, for the binary classification of bio-activity and toxicity, we studied the relationship between predicted probability and uncertainties.
    Our analysis shows that prediction with a lower uncertainty turned out to be more accurate, indicating that the uncertainty can be regarded as the confidence of prediction.  
\end{itemize}

\section{Theoretical backgrounds}
\subsection{Bayesian neural network}

For a given training set $\{ \textbf{X}, \textbf{Y} \}$, let  $p(\textbf{Y}|\textbf{X},\textbf{w})$ and $ p(\textbf{w})$ be a model likelihood and a prior distribution for a parameter $\textbf{w} \in \Omega$, respectively. 
Under the Bayesian framework, the model parameter and output are considered as random variables. 
The posterior distribution is given by
\begin{equation} \label{eq:1}
p(\textbf{w}|\textbf{X},\textbf{Y}) = \frac{p(\textbf{Y}|\textbf{X},\textbf{w})p(\textbf{w})}{p(\textbf{Y}|\textbf{X})}
\end{equation}
and the predictive distribution is defined as
\begin{equation} \label{eq:2}
p(\textbf{y}^*|\textbf{x}^*,\textbf{X},\textbf{Y}) = \int _\Omega p(\textbf{y}^*|\textbf{x}^*,\textbf{w}) p(\textbf{w}|\textbf{X},\textbf{Y}) d\textbf{w}
\end{equation}
for a new input $\textbf{x}^*$ and an output $\textbf{y}^*$. 
These simple formulations make the two following tasks possible: (i) assessing uncertainty of the random variables in a conditional manner and (ii) predicting a distribution of the new output $\textbf{y}^*$ given both the new input $\textbf{x}^*$ and the training set $\{ \textbf{X}, \textbf{Y} \}$.

However, direct computation of eq. \eqref{eq:2} is often infeasible when deep neural network models are exploited because the integration over the whole parameter space $\Omega$ entails heavy computational costs.
Many practical approximation methods have been proposed to handle this computation cost.
A variational inference, one of the most popular approximation methods, approximates the posterior distribution with a tractable distribution $q_{\theta}(\textbf{w})$ parametrized by a variational parameter $\theta$ \cite{blundell2015weight, graves2011practical}.
Minimizing the Kullback-Leibler divergence, 
\begin{equation} \label{eq:3}
\textrm{KL}(q_{\theta}(\textbf{w}) \Vert p(\textbf{w}|\textbf{X},\textbf{Y})) = \int _\Omega q_{\theta}(\textbf{w}) \log \frac{q_{\theta}(\textbf{w})}{p(\textbf{w}|\textbf{X},\textbf{Y})} d\textbf{w},
\end{equation}
makes the two distributions similar to one another in principle.
We can replace the intractable posterior distribution in \eqref{eq:3} with $p(\textbf{Y}|\textbf{X},\textbf{w})p(\textbf{w})$ due to the Bayes' theorem \eqref{eq:1}.
Then, our minimization objective, called the negative evidence lower-bound, is
\begin{equation} \label{eq:4}
\mathcal{L}_{\textrm{VI}}(\theta) = -\int _\Omega q_{\theta}(\textbf{w})\log{p(\textbf{Y}|\textbf{X},\textbf{w})}d\textbf{w}+\textrm{KL}(q_{\theta}(\textbf{w}) \Vert p(\textbf{w})).
\end{equation}

In order to implement Bayesian models, we need to be cautious in choosing a variational distribution $q_{\theta}(\textbf{w})$. 
\citeauthor{blundell2015weight} proposed to use a product of Gaussian distributions for the variational distribution $q_{\theta}(\textbf{w})$.
In addition, a multiplicative normalizing flow \cite{louizos2017multiplicative} can be applied to increase the expressive power of variational distribution.
However, the two approaches often require a large number of weight parameters. 
The Monte-Carlo dropout (MC-dropout) using a dropout\cite{srivastava2014dropout} variational distribution approximates the posterior distribution by a product of Bernoulli distribution \cite{gal2016dropout}.
The MC-dropout is practical in that it does not need extra learnable parameters to model the variational posterior distribution and the integration over the whole parameter space can be easily approximated with the summation of models sampled by a Monte-Carlo estimator \cite{gal2016uncertainty, gal2016dropout}.
Thus, we adopted the MC-dropout in this work. 

\subsection{Uncertainty quantification with Bayesian neural network}
A variational inference approximating a posterior with a variational distribution $q_{\theta}(\textbf{w})$ provides a variational predictive distribution of a new output $\textbf{y}^*$ given a new input $\textbf{x}^*$ as 
\begin{equation} \label{eq:5}
q_{\theta}^*(\textbf{y}^*|\textbf{x}^*) = \int _\Omega q_{\theta}(\textbf{w}) p(\textbf{y}^*|f^{\textbf{w}}(\textbf{x}^*)) d\textbf{w},
\end{equation}
where $f^{\textbf{w}}(\textbf{x}^*)$ is a model output with a given $\textbf{w}$.
For regression tasks, a predictive mean of this distribution with $T$ times of MC sampling is estimated by
\begin{equation} \label{eq:6}
    \hat{E} [\textbf{y}^*|\textbf{x}^*] = \frac{1}{T}\sum_{t=1}^{T} f^{\hat{\textbf{w}}_t}(\textbf{x}^*),
\end{equation}
and a predictive variance is estimated by
\begin{equation} \label{eq:7}
    \widehat{Var} [\textbf{y}^*|\textbf{x}^*] = \sigma^2I + \frac{1}{T} \sum_{t=1}^{T} f^{\hat{\textbf{w}}_t}(\textbf{x}^*)^T f^{\hat{\textbf{w}}_t}(\textbf{x}^*) - \hat{E} [\textbf{y}^* | \textbf{x}^*]^T  \hat{E} [\textbf{y}^* | \textbf{x}^*],
\end{equation}
with $\hat{\textbf{w}}_t$ drawn from $q_{\theta}(\textbf{w})$ at the sampling step $t$ and an assumption $p(\textbf{y}^*|f^{\textbf{w}}(\textbf{x}^*)) = N(\textbf{y}^*;f^{\textbf{w}}(\textbf{x}^*), \sigma^2I)$. Here, the model assumes a homoscedasticity with a known quantity, meaning that every data point gives a distribution with a same variance $\sigma^2$.
Further to this, obtaining the distributions with different variances allows deducing a heteroscedastic uncertainty.
Assuming the heteroscedasticity, the output given the $t$-th sample $\hat{\textbf{w}}_t$ is 
\begin{equation} \label{eq:8}
    [\hat{\textbf{y}}_{t}^*, \hat{\sigma}_{t}] = f^{\hat{\textbf{w}}_t}(\textbf{x}^*).
\end{equation}
The heteroscedastic predictive uncertainty given by \eqref{eq:9} can be partitioned into two different uncertainties: aleatoric and epistemic uncertainties.
\begin{equation} \label{eq:9} 
    \widehat{Var} [\textbf{y}^*|\textbf{x}^*] = \underbrace{\frac{1}{T} \sum_{t=1}^{T} (\hat{\textbf{y}}_{t}^{*})^2 - (\frac{1}{T} \sum_{t=1}^{T} \hat{\textbf{y}}_{t}^*)^2}_\textrm{epistemic} + \underbrace{\frac{1}{T} \sum_{t=1}^{T} \hat{\sigma}_{t}^2}_\textrm{aleatoric}.
\end{equation}
The aleatoric uncertainty arises from data inherent noise, while the epistemic uncertainty is related to the model incompleteness. 
Note that the latter can be reduced by increasing the amount of training data, because it comes from insufficient amount of data as well as the use of inappropriate model \cite{der2009aleatory}.

In classification problems, \citeauthor{kwon2018uncertainty} proposed a natural way to quantify aleatoric and epistemic uncertainties as follows.
\begin{equation} \label{eq:10} 
\widehat{Var} [\textbf{y}^* | \textbf{x}^*] = \underbrace{\frac{1}{T} \sum_{t=1}^{T}(\hat{\textbf{y}}_t^* - \bar{\textbf{y}})(\hat{\textbf{y}}_t^* - \bar{\textbf{y}})^T}_\textrm{epistemic} + \underbrace{\frac{1}{T} \sum_{t=1}^{T} (\textrm{diag}(\hat{\textbf{y}}_t^*) - (\hat{\textbf{y}}_t^*)(\hat{\textbf{y}}_t^*)^T)}_\textrm{aleatoric}, 
\end{equation}
where $\bar{\textbf{y}} = \sum_{t=1}^{T} \hat{\textbf{y}}_t^*/T$ and $\hat{\textbf{y}}_t^* = \textrm{softmax}(\textbf{f}^{\hat{\textbf{w}}_t}(\textbf{x}^*))$.
While \citeauthor{kendall2017uncertainties}'s method requires extra parameters $\hat{{\sigma}}_t$ at the last hidden layer and often causes unstable parameter updates in a training phase,\cite{kendall2017uncertainties} the method in \citeauthor{kwon2018uncertainty} has advantages in that models do not need the extra parameters.\cite{kwon2018uncertainty}
The equation \eqref{eq:10} also utilizes a functional relationship between mean and variance of multinomial random variables.
We refer to \citeauthor{kwon2018uncertainty} for more details.

\subsection{Graph convolutional network for molecular property predictions}

Molecules, social graphs, images and language sentences can be represented as graph structures \cite{battaglia2018relational}. 
GCN is one of the most popular graph neural networks and is widely adopted to process molcular graphs. 
Inputs to the GCN is $\textbf{G}=(\textbf{A},\textbf{X})$, 
where $\textbf{A} \in \mathds{R}^{N \times N}$ is an adjacency matrix with the number of nodes $N$ and $\textbf{X} = \textbf{H}^{(0)}\in \mathds{R}^{N \times F_{inp}}$ is a set of initial node features whose dimensionality is $F_{inp}$.  
The GCN gives new node features as follows.
\begin{equation} \label{eq:11}
\textbf{H}^{(l+1)} = \textrm{ReLU}(\textbf{A}\textbf{H}^{(l)}\textbf{W}^{(l)}),
\end{equation}
where $\textbf{H}^{(l)} \in \mathds{R}^{N \times F}$ and $\textbf{W}^{(l)} \in \mathds{R}^{F \times F}$ are node features and weight parameters for the $l$-th graph convolution layer for $l \in \{ 0, \dots, L-1\}$, respectively.
The GCN updates node features $\textbf{H}^{(l+1)}$ with information of only adjacent nodes.

Applying a self-attention\cite{vaswani2017attention} enables the GCN to learn relations between node pairs by reflecting the importance of adjacent nodes.\cite{velickovic2017graph} 
Updating node features with the $K$-head self-attention is given by 
\begin{equation} \label{eqn:eq12} 
\tilde{\textbf{H}}_i^{(l+1)} =  [\text{ReLU}(\sum_{j \in \mathcal{N}_i} \alpha_{ij,1}^{(l)} \textbf{H}_{j}^{(l)} \textbf{W}_{1}^{(l)}), ..., \text{ReLU}(\sum_{j \in \mathcal{N}_i} \alpha_{ij,K}^{(l)} \textbf{H}_{j}^{(l)} \textbf{W}_{K}^{(l)})] \textbf{W}_{O}^{(l)},
\end{equation}
where $\mathcal{N}_i$ denotes the adjacent nodes of the $i$-th node, 
$H_j^{(l)} \in \mathds{R}^{1 \times F}$ is the $j$-th node feature updated at $l$-th graph convolution,
$W_k^{(l)} \in \mathds{R}^{F \times F}$ is a weight parameter for the $k$-th attention head, 
$\textbf{W}_O^{(l)} \in \mathds{R}^{K F \times F}$ is a weight parameter to combine the node features from $K$-different attention heads, 
and the attention coefficient $\alpha_{ij,k}^{(l)}$ is given by
\begin{equation}
\alpha_{ij,k}^{(l)} = \text{tanh}((\textbf{H}_i^{(l)} \textbf{W}_k^{(l)}) \textbf{C}_k^{(l)} (\textbf{H}_j^{(l)} \textbf{W}_k^{(l)})^T),
\end{equation}
where $\textbf{C}_k^{(l)} \in \mathds{R}^{F \times F}$ is a weight parameter.  

In addition, the GCN has room for improvement because its accuracy is gradually lowered as the number of graph convolution layers increases. \cite{kipf2016semi, ryu2018deeply} 
We used a gated-skip connection to prevent this problem as follows.
\begin{equation} \label{eq:13}
\textbf{H}^{(l+1)} = \textbf{r} \odot \tilde{\textbf{H}}^{(l+1)} + (\textbf{1}-\textbf{r}) \odot \textbf{H}^{(l)}, \quad \textbf{r} = \textrm{sigmoid}(\textbf{U}_{r,1} \textbf{H}^{(l)} + \textbf{U}_{r,2} \tilde{\textbf{H}}^{(l+1)} + \textbf{b}_r),
\end{equation}
where $\textbf{U}_{r,1}$ and $\textbf{U}_{r,2}$ are trainable parameters and $\odot$ denotes Hadamard product. 

After computing the node features $L$-times by following eq. \eqref{eq:13}, a graph feature $\textbf{z}_{\textrm{G}} \in \mathds{R}^{d_G}$ is aggregated as the summation of all node features in a set of nodes $\mathcal{V}$, 
\begin{equation} \label{eq:14}
\textbf{z}_{\textrm{G}} = \sum_{v \in \mathcal{V}} \textrm{MLP}_1(\textbf{H}_v^{(L)}),   
\end{equation}
where $\textrm{MLP}$ denotes a multi-layer perceptron. The graph feature is invariant to permutations of the node states. A molecular property, which is the final output from the model, is a function of the graph feature.
\begin{equation} \label{eq:15}
y_{pred} = \textrm{MLP}_2(\textbf{z}_{\textrm{G}}).
\end{equation}

\section{Implementation details}
\subsection{Model architecture}
\begin{figure}
    \centering
    \includegraphics[width=0.5\textwidth]{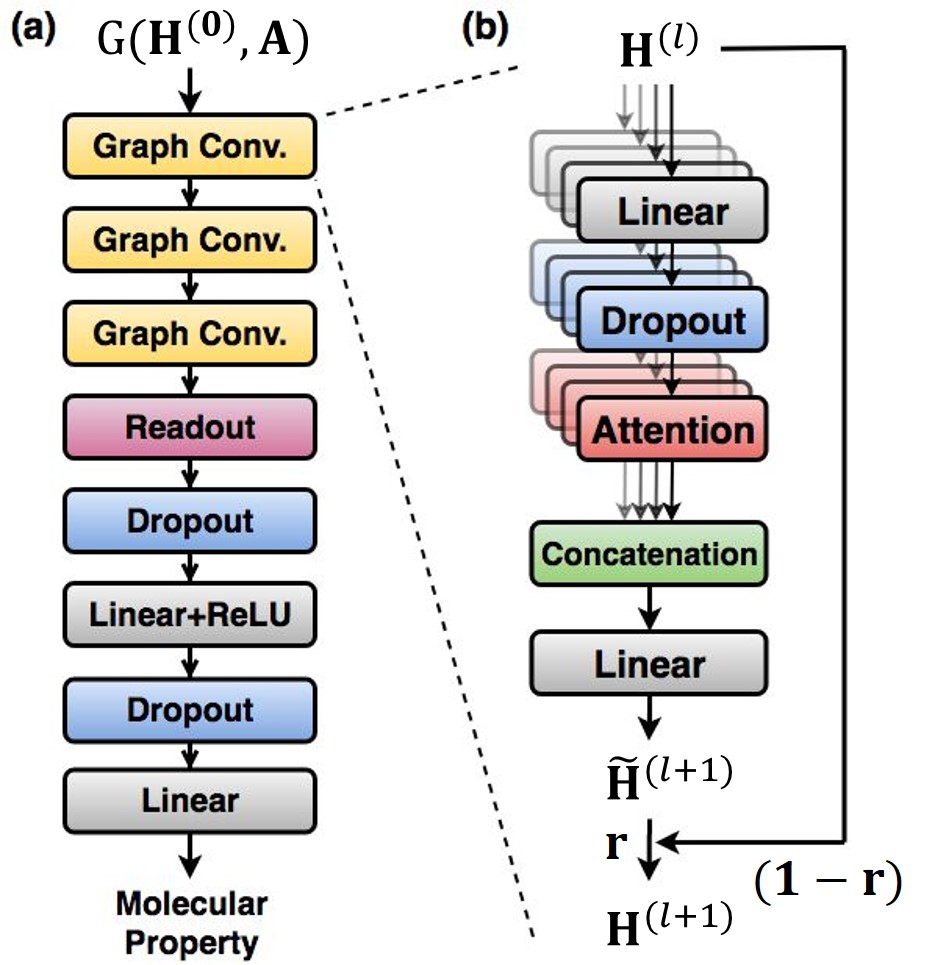}
    \caption{The architecture of Bayesian GCN used in this work. (a) The entire model is composed of three augemented graph convolutional layers, readout layers and three linear layers with non-linear activation. (b) Detailed description of the graph convolution layer augmented with attention and gate mechanisms. We added dropout layers in order for the model parameters to have stochasticity.}
    \label{fig:model_architecture}
\end{figure}

As illustrated in Figure \ref{fig:model_architecture}, our graph convolutional MC-dropout network used in this work consists of the following three parts:
\begin{itemize}
    \item Three augmented graph convolution layers update node features according to \eqref{eq:13}. The number of self-attention head is four. The dimension of output from each layer is ($N \times F$) = ($75\times32$). 
    \item A readout function produces a graph feature whose dimension $d_G$ is 256 by following \eqref{eq:14}. 
    \item A feed-forward MLP, which is composed of two fully-connected layers, turns out a molecular property. The hidden dimension of each fully-connected layer is 256.
\end{itemize}
In order for the model parameters to have stochasticity, we applied dropouts at every hidden layer. 
Note that we did not use the standard dropout with a pre-defined dropout rate, but used Concrete dropout\cite{gal2017concrete} to develop as an accurate Bayesian model as possible. 
By using the Concrete dropout, we can obtain an optimal dropout rate for individual hidden layer by a stochastic optimization. 
We used Gaussian priors $\mathcal{N}(0, l^2)$ with length scale $l=10^{-4}$ for all model parameters. 
In the training phase, we used the Adam optimizer\cite{kingma2014adam} with an initial learning rate $10^{-3}$, and the learning rate is decayed by half at every 10 epoch.
The number of total training epoches is $100$ and the batch size is $100$. 
We randomly split datasets in the ratio of $(0.72:0.08:0.2)$ for training, validation and test. 
The code used for the experiments is available at \url{https://github.com/seongokryu/uq-molecule}.

\section{Experiments}

\subsection{Implication of data quality on aleatoric and epistemic uncertainties}
\begin{figure}
    \centering
    \includegraphics[width=1.0\textwidth]{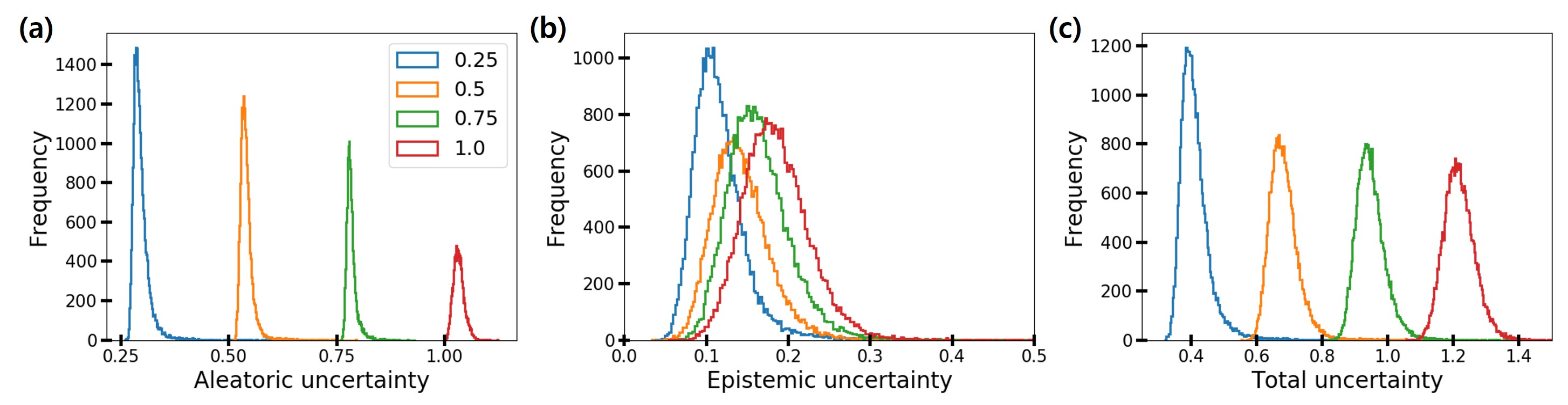}
    \caption{Histograms of (a) aleatoric , (b) epistemic and (c) total uncertanties as the amount of additive noise $\sigma^2$ increases.}
    \label{fig:Bayesian_logP}
\end{figure}

In this experiment, we applied the uncertainty quantification method to a simple example, logP prediction.
We chose this example because we can obtain the logP value of molecules from the analytic expression of logP as implemented in the RDKit \cite{landrum2006rdkit} without data inherent noise. 
To examine the effect of data quality on uncertainties, we adjust the extent of noise in logP by adding a random Gaussian noise $\epsilon \sim \mathcal{N}(0,\sigma^2)$. 
We trained the model with 97,287 samples and analyzed uncertainties of each predicted logP for 27,023 samples. The samples were chosen randomly from the ZINC dataset.

Figure \ref{fig:Bayesian_logP} shows the distribution of the three uncertainties as a function of the amount of additive noise $\sigma^2$. 
As the noise level increases, the aleatoric and total uncertainties increase, but the epistemic uncertainty is slightly changed. 
This result verifies that the aleatoric uncertainty arises from data inherent noises, while the epistemic uncertainty does not depend on data quality. 
Theoretically, the epistemic uncertainty should not increase by the changes in the amount of data noise.
We guess that the slight change of the epistemic uncertainty arises from the stochastic numerical optimization of model parameters. 

\subsection{Evaluating quality of synthetic data based on uncertainty analysis}

\begin{figure}
    \centering
    \includegraphics[width=0.5\textwidth]{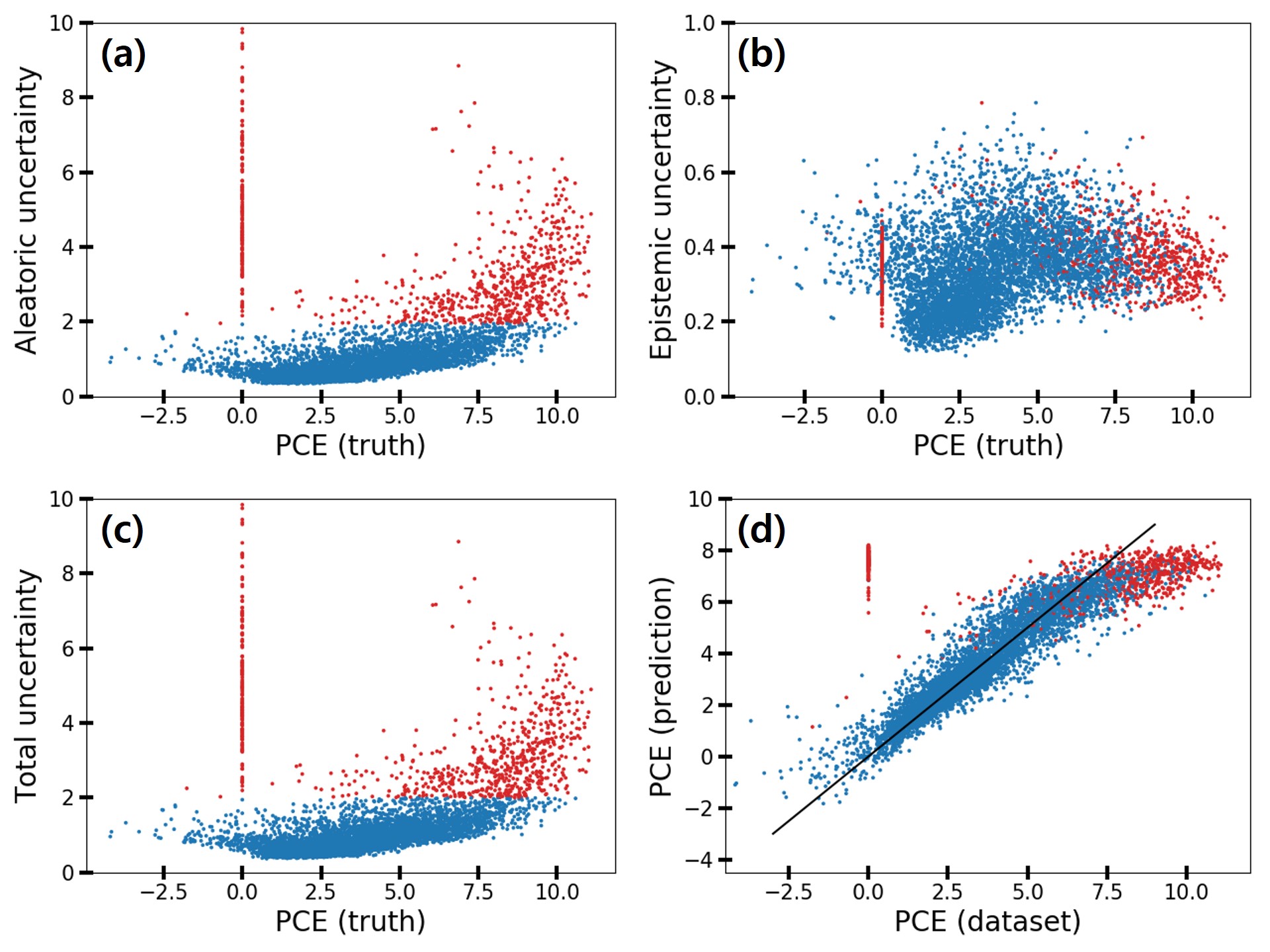}
    \caption{(a) Aleatoric, (b) epistemic, (c) total uncertainties and (d) predicted PCE against the PCE value in the dataset. The samples colored in red show the total uncertainty greater than two.}
    \label{fig:Bayesian_CEP}
\end{figure}

Based on the analysis of the previous experiment, we attempted to evaluate the quality of synthetic data.
Synthetic PCE values in the CEP dataset \cite{hachmann2011harvard} was obtained from the Scharber model with statistical approximations \cite{scharber2006design}.
In this procedure, unintentional errors can be involved in the resulting synthetic data.
Since the aleatoric uncertainty arises due to data quality, we evaluated quality of the synthetic data by analyzing the uncertainties of predicted PCE values.
We used the same dataset in \citeauthor{duvenaud2015convolutional} \footnote{\url{https://github.com/HIPS/neural-fingerprint}} for training and test.

Figure \ref{fig:Bayesian_CEP} shows the scatter plot of three uncertainties in the CEP predictions for 5,995 molecules in the test set.
Samples with the total uncertainty greater than two are highlighted with red color. 
Some samples with large PCE values above eight had relatively large total uncertainties. 
Their PCE values deviated considerably from the black line in Figure \ref{fig:Bayesian_CEP}-(d). 
More interestingly, we found that most molecules with the zero PCE value had large total uncertainties as well. 
Those large uncertainties came from the aleatoric uncertainty as depicted in Figure \ref{fig:Bayesian_CEP}-(a), indicating that the data quality of those particular samples is relatively poor. 
Hence, we speculated that data inherent noises might cause large prediction errors.

To elaborate the origin of such errors, we investigated the procedure of obtaining the PCE values. The Havard Organic Photovolatic Dataset \cite{lopez2016harvard} contains both experimental and synthetic PCE values of 350 organic photovoltaic materials.
The synthetic PCE values were computed according to \eqref{eq:16}, which is the result of the Scharber model \cite{scharber2006design}. 
\begin{equation}\label{eq:16}
\textrm{PCE} \propto V_{OC} \times FF \times J_{SC},
\end{equation}
where $V_{OC}$ is an open circuit potential, $FF$ is a fill factor, and $J_{SC}$ is a short circuit current density.
$FF$ was set to 65\%. $V_{OC}$ and $J_{SC}$ were obtained from electronic structure calculations of molecules.\cite{hachmann2011harvard}
We found that $J_{SC}$ of some molecules were zero or nearly zero, resulting in zero or almost zero synthetic PCE values, in contrast to their non-zero experimental PCE values.
Especially, $J_{SC}$ and PCE values computed using the M06-2X functional \cite{zhao2008m06} were almost zero consistently.
We suspect that those approximated values caused a significant drop of data quality, resulting in large aleatoric uncertainties as highlighted in Figure \ref{fig:Bayesian_CEP}. 
Consequently, the data noise due to poorly fabricated data was identified as the large aleatoric uncertainties.

\subsection{Uncertainty as confidence indicator: bio-activity and toxicity classification}

\begin{figure}
    \centering
    \includegraphics[width=1.0\textwidth]{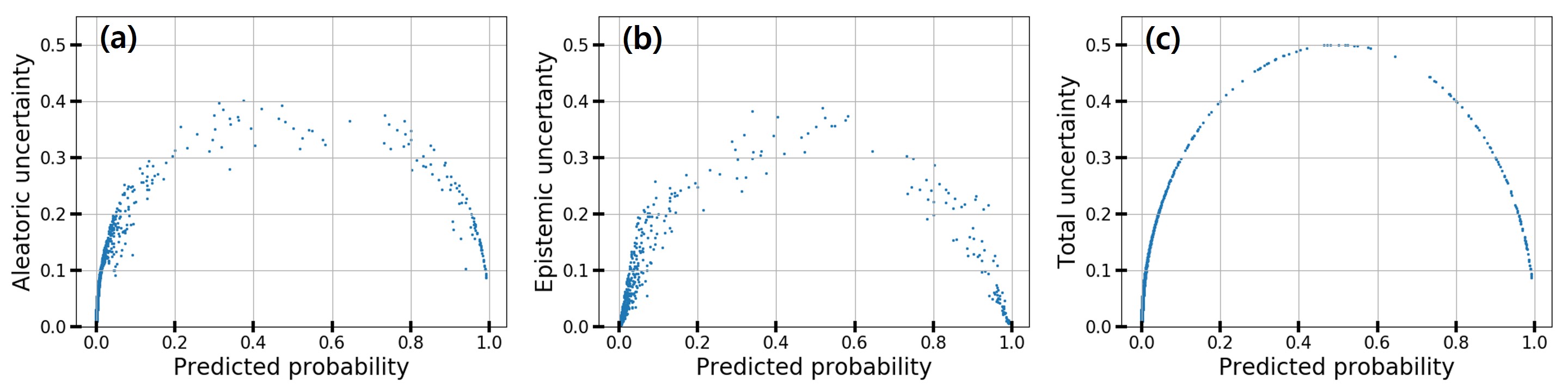}
    \caption{(a) Aleatoric, (b) epistemic and (c) total uncertainty of predicted probabilities in the classification of bio-activity against the EGFR target.}
    \label{fig:6}
\end{figure}


In this experiment, we demonstrate that the uncertainty analysis can lead reliable classification.
In classification problems, it tends to interpret the final outputs from a sigmoid or softmax activation as their confidence, which means that the higher the output probability, the higher the prediction accuracy.
However, as \citeauthor{gal2016dropout} pointed out, such interpretation is erroneous. \cite{gal2016dropout}
Thus, we applied the uncertainty quantification on the bio-activity and toxicity classification problems and show that the predictive uncertainty can be used as the confidence of outcomes.

\begin{figure}
    \centering
    \includegraphics[width=1.0\textwidth]{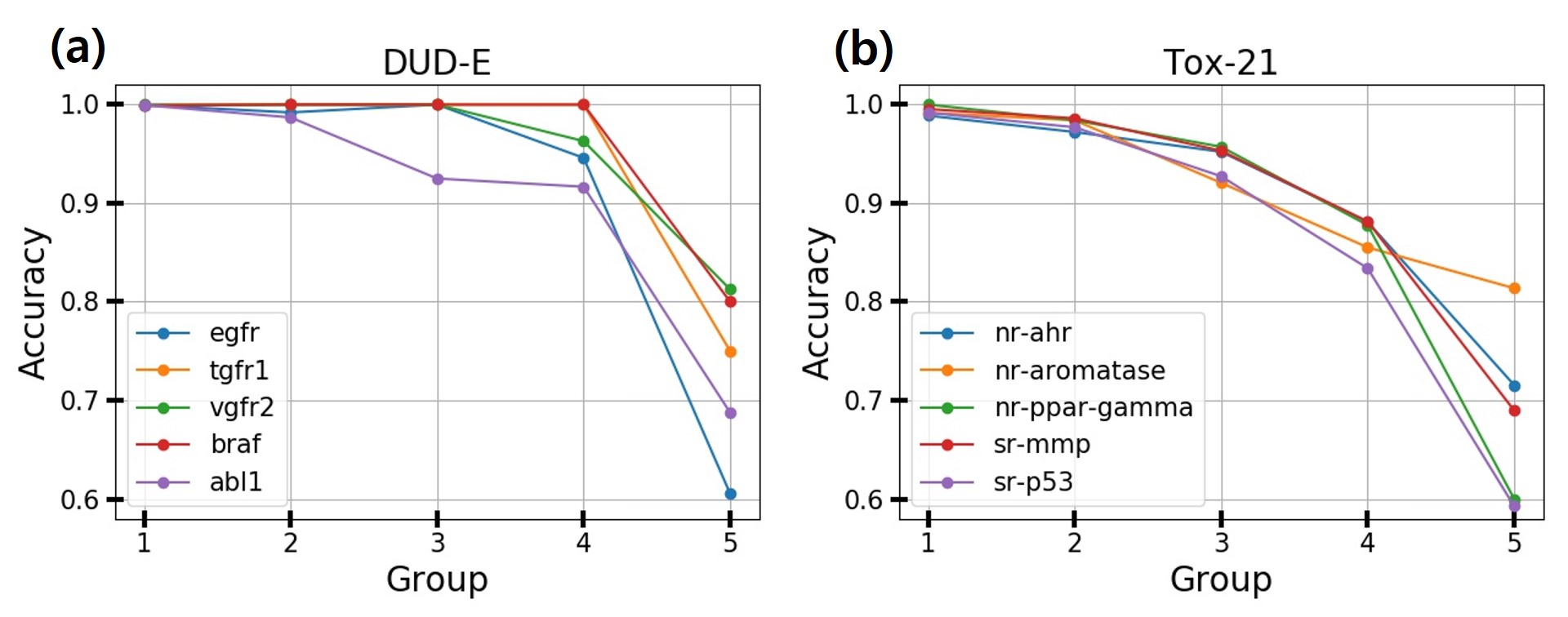}
    \caption{Test accuracy for the classifications of (a) bio-activities against the five target proteins in the DUD-E set and (b) the five toxic effects in the Tox21 set.}
    \label{fig:Bayesian_group_accuracy}
\end{figure}

We trained the Bayesian GCN using 25,627 molecules with the labels for EGFR-activity in the DUD-E dataset.
Figure \ref{fig:6} shows the results for 7,118 molecules in the test set. 
In order for the predictive uncertainty to be interpreted as a confidence, its value should be minimum on the output probability of zero or one and should be maximum on that of 0.5. 
Indeed, the total uncertainty predicted from our model shows such behaviour. 
In other words, more uncertain outcomes have lower predictive probability values. 
We also noted that the aleatoric uncertainty affected the total uncertainty more significantly than the epistemic uncertainty did.

To further investigate a relationship between accuracy and uncertainty, we trained the Bayesian GCN for various bio-activity labels in the DUD-E dataset and toxicity labels in the Tox21 dataset. 
Then, we sorted the molecules in the order of increasing uncertainty and then divided them into five groups as follows: 
molecules in the $i$-th group have total uncertainties in the range of $((i-1)\times0.1, i\times0.1)$.
Figure \ref{fig:Bayesian_group_accuracy} shows the classification accuracy of each group; (a) and (b) denote the classification results of bio-acitvities against the five different targets and the five different toxicities of Tox21 set molecules, respectively. 
This result is an evidence that the uncertainty can be used as a confidence indicator in binary classification problems.

\section{Conclusion}
Deep neural network models show promising performances in the prediction of molecular properties. 
In practical applications, however, a lack of data quality and quantity discourages developing accurate models. 
To make reliable decisions in such a case, we have proposed to analyze uncertainties in the prediction results by using the Bayesian GCN.

Our first experiment on the logP prediction showed that data inherent noise can be identified by the aleatoric uncertainty. 
The aleatoric uncertainty in the predicted logP values increases as the amount of noise increases. 
In contrast, the epistemic uncertainty slightly depends on the data noise as expected. 
In the second experiment, we applied the uncertainty analysis to the Harvard Clean Energy Project dataset. 
It was able to identify erroneous data by noting the abnormally increased aleatoric uncertainty in the poorly approximated synthetic data, which is helpful to find the source of the errors.
In the third experiment of bio-activity and toxicity predictions, we showed that the uncertainty is closely related to the confidence of prediction for binary classification problems. 
As grouping the molecules in the increasing order of uncertainty, the groups with lower uncertainty show higher accuracy than those with higher uncertainty.

We have demonstrated how useful the uncertainty quantification is in molecular applications.
By using the Bayesian GCN, we can analyze the quality of data that is often noisy because of the stochastic nature of experimental results.  
From the relationship between output probability and confidence of prediction, it is able to extract more reliable results selectively from entire predictions, 
which is critical to making a desirable decision.
Such analysis can be used to screen bio-active and toxic molecules, where reliable prediction is vital.
We believe that our study on the uncertainty quantification of molecular properties offers insights to tackle AI-safety problems in molecular applications. 

\bibliography{achemso-demo}

\providecommand{\latin}[1]{#1}
\makeatletter
\providecommand{\doi}
  {\begingroup\let\do\@makeother\dospecials
  \catcode`\{=1 \catcode`\}=2\doi@aux}
\providecommand{\doi@aux}[1]{\endgroup\texttt{#1}}
\makeatother
\providecommand*\mcitethebibliography{\thebibliography}
\csname @ifundefined\endcsname{endmcitethebibliography}
  {\let\endmcitethebibliography\endthebibliography}{}
\begin{mcitethebibliography}{49}
\providecommand*\natexlab[1]{#1}
\providecommand*\mciteSetBstSublistMode[1]{}
\providecommand*\mciteSetBstMaxWidthForm[2]{}
\providecommand*\mciteBstWouldAddEndPuncttrue
  {\def\EndOfBibitem{\unskip.}}
\providecommand*\mciteBstWouldAddEndPunctfalse
  {\let\EndOfBibitem\relax}
\providecommand*\mciteSetBstMidEndSepPunct[3]{}
\providecommand*\mciteSetBstSublistLabelBeginEnd[3]{}
\providecommand*\EndOfBibitem{}
\mciteSetBstSublistMode{f}
\mciteSetBstMaxWidthForm{subitem}{(\alph{mcitesubitemcount})}
\mciteSetBstSublistLabelBeginEnd
  {\mcitemaxwidthsubitemform\space}
  {\relax}
  {\relax}

\bibitem[Gomes \latin{et~al.}(2017)Gomes, Ramsundar, Feinberg, and
  Pande]{gomes2017atomic}
Gomes,~J.; Ramsundar,~B.; Feinberg,~E.~N.; Pande,~V.~S. Atomic convolutional
  networks for predicting protein-ligand binding affinity. \emph{arXiv preprint
  arXiv:1703.10603} \textbf{2017}, \relax
\mciteBstWouldAddEndPunctfalse
\mciteSetBstMidEndSepPunct{\mcitedefaultmidpunct}
{}{\mcitedefaultseppunct}\relax
\EndOfBibitem
\bibitem[Jim{\'e}nez \latin{et~al.}(2018)Jim{\'e}nez, Skalic,
  Mart{\'\i}nez-Rosell, and De~Fabritiis]{jimenez2018k}
Jim{\'e}nez,~J.; Skalic,~M.; Mart{\'\i}nez-Rosell,~G.; De~Fabritiis,~G. K DEEP:
  Protein--Ligand Absolute Binding Affinity Prediction via 3D-Convolutional
  Neural Networks. \emph{Journal of chemical information and modeling}
  \textbf{2018}, \emph{58}, 287--296\relax
\mciteBstWouldAddEndPuncttrue
\mciteSetBstMidEndSepPunct{\mcitedefaultmidpunct}
{\mcitedefaultendpunct}{\mcitedefaultseppunct}\relax
\EndOfBibitem
\bibitem[Mayr \latin{et~al.}(2016)Mayr, Klambauer, Unterthiner, and
  Hochreiter]{mayr2016deeptox}
Mayr,~A.; Klambauer,~G.; Unterthiner,~T.; Hochreiter,~S. DeepTox: toxicity
  prediction using deep learning. \emph{Frontiers in Environmental Science}
  \textbf{2016}, \emph{3}, 80\relax
\mciteBstWouldAddEndPuncttrue
\mciteSetBstMidEndSepPunct{\mcitedefaultmidpunct}
{\mcitedefaultendpunct}{\mcitedefaultseppunct}\relax
\EndOfBibitem
\bibitem[{\"O}zt{\"u}rk \latin{et~al.}(2018){\"O}zt{\"u}rk, {\"O}zg{\"u}r, and
  Ozkirimli]{ozturk2018deepdta}
{\"O}zt{\"u}rk,~H.; {\"O}zg{\"u}r,~A.; Ozkirimli,~E. DeepDTA: deep drug--target
  binding affinity prediction. \emph{Bioinformatics} \textbf{2018}, \emph{34},
  i821--i829\relax
\mciteBstWouldAddEndPuncttrue
\mciteSetBstMidEndSepPunct{\mcitedefaultmidpunct}
{\mcitedefaultendpunct}{\mcitedefaultseppunct}\relax
\EndOfBibitem
\bibitem[De~Cao and Kipf(2018)De~Cao, and Kipf]{de2018molgan}
De~Cao,~N.; Kipf,~T. MolGAN: An implicit generative model for small molecular
  graphs. \emph{arXiv preprint arXiv:1805.11973} \textbf{2018}, \relax
\mciteBstWouldAddEndPunctfalse
\mciteSetBstMidEndSepPunct{\mcitedefaultmidpunct}
{}{\mcitedefaultseppunct}\relax
\EndOfBibitem
\bibitem[G{\'o}mez-Bombarelli \latin{et~al.}(2018)G{\'o}mez-Bombarelli, Wei,
  Duvenaud, Hern{\'a}ndez-Lobato, S{\'a}nchez-Lengeling, Sheberla,
  Aguilera-Iparraguirre, Hirzel, Adams, and Aspuru-Guzik]{gomez2018automatic}
G{\'o}mez-Bombarelli,~R.; Wei,~J.~N.; Duvenaud,~D.;
  Hern{\'a}ndez-Lobato,~J.~M.; S{\'a}nchez-Lengeling,~B. \latin{et~al.}
  Automatic chemical design using a data-driven continuous representation of
  molecules. \emph{ACS central science} \textbf{2018}, \emph{4}, 268--276\relax
\mciteBstWouldAddEndPuncttrue
\mciteSetBstMidEndSepPunct{\mcitedefaultmidpunct}
{\mcitedefaultendpunct}{\mcitedefaultseppunct}\relax
\EndOfBibitem
\bibitem[Guimaraes \latin{et~al.}(2017)Guimaraes, Sanchez-Lengeling, Outeiral,
  Farias, and Aspuru-Guzik]{guimaraes2017objective}
Guimaraes,~G.~L.; Sanchez-Lengeling,~B.; Outeiral,~C.; Farias,~P. L.~C.;
  Aspuru-Guzik,~A. Objective-reinforced generative adversarial networks (ORGAN)
  for sequence generation models. \emph{arXiv preprint arXiv:1705.10843}
  \textbf{2017}, \relax
\mciteBstWouldAddEndPunctfalse
\mciteSetBstMidEndSepPunct{\mcitedefaultmidpunct}
{}{\mcitedefaultseppunct}\relax
\EndOfBibitem
\bibitem[Jin \latin{et~al.}(2018)Jin, Barzilay, and Jaakkola]{jin2018junction}
Jin,~W.; Barzilay,~R.; Jaakkola,~T. Junction Tree Variational Autoencoder for
  Molecular Graph Generation. \emph{arXiv preprint arXiv:1802.04364}
  \textbf{2018}, \relax
\mciteBstWouldAddEndPunctfalse
\mciteSetBstMidEndSepPunct{\mcitedefaultmidpunct}
{}{\mcitedefaultseppunct}\relax
\EndOfBibitem
\bibitem[Kusner \latin{et~al.}(2017)Kusner, Paige, and
  Hern{\'a}ndez-Lobato]{kusner2017grammar}
Kusner,~M.~J.; Paige,~B.; Hern{\'a}ndez-Lobato,~J.~M. Grammar variational
  autoencoder. \emph{arXiv preprint arXiv:1703.01925} \textbf{2017}, \relax
\mciteBstWouldAddEndPunctfalse
\mciteSetBstMidEndSepPunct{\mcitedefaultmidpunct}
{}{\mcitedefaultseppunct}\relax
\EndOfBibitem
\bibitem[Li \latin{et~al.}(2018)Li, Vinyals, Dyer, Pascanu, and
  Battaglia]{li2018learning}
Li,~Y.; Vinyals,~O.; Dyer,~C.; Pascanu,~R.; Battaglia,~P. Learning deep
  generative models of graphs. \emph{arXiv preprint arXiv:1803.03324}
  \textbf{2018}, \relax
\mciteBstWouldAddEndPunctfalse
\mciteSetBstMidEndSepPunct{\mcitedefaultmidpunct}
{}{\mcitedefaultseppunct}\relax
\EndOfBibitem
\bibitem[Segler \latin{et~al.}(2017)Segler, Kogej, Tyrchan, and
  Waller]{segler2017generating}
Segler,~M.~H.; Kogej,~T.; Tyrchan,~C.; Waller,~M.~P. Generating focused
  molecule libraries for drug discovery with recurrent neural networks.
  \emph{ACS central science} \textbf{2017}, \emph{4}, 120--131\relax
\mciteBstWouldAddEndPuncttrue
\mciteSetBstMidEndSepPunct{\mcitedefaultmidpunct}
{\mcitedefaultendpunct}{\mcitedefaultseppunct}\relax
\EndOfBibitem
\bibitem[You \latin{et~al.}(2018)You, Liu, Ying, Pande, and
  Leskovec]{you2018graph}
You,~J.; Liu,~B.; Ying,~R.; Pande,~V.; Leskovec,~J. Graph Convolutional Policy
  Network for Goal-Directed Molecular Graph Generation. \emph{arXiv preprint
  arXiv:1806.02473} \textbf{2018}, \relax
\mciteBstWouldAddEndPunctfalse
\mciteSetBstMidEndSepPunct{\mcitedefaultmidpunct}
{}{\mcitedefaultseppunct}\relax
\EndOfBibitem
\bibitem[Segler \latin{et~al.}(2018)Segler, Preuss, and
  Waller]{segler2018planning}
Segler,~M.~H.; Preuss,~M.; Waller,~M.~P. Planning chemical syntheses with deep
  neural networks and symbolic AI. \emph{Nature} \textbf{2018}, \emph{555},
  604\relax
\mciteBstWouldAddEndPuncttrue
\mciteSetBstMidEndSepPunct{\mcitedefaultmidpunct}
{\mcitedefaultendpunct}{\mcitedefaultseppunct}\relax
\EndOfBibitem
\bibitem[Wei \latin{et~al.}(2016)Wei, Duvenaud, and
  Aspuru-Guzik]{wei2016neural}
Wei,~J.~N.; Duvenaud,~D.; Aspuru-Guzik,~A. Neural networks for the prediction
  of organic chemistry reactions. \emph{ACS central science} \textbf{2016},
  \emph{2}, 725--732\relax
\mciteBstWouldAddEndPuncttrue
\mciteSetBstMidEndSepPunct{\mcitedefaultmidpunct}
{\mcitedefaultendpunct}{\mcitedefaultseppunct}\relax
\EndOfBibitem
\bibitem[Zhou \latin{et~al.}(2017)Zhou, Li, and Zare]{zhou2017optimizing}
Zhou,~Z.; Li,~X.; Zare,~R.~N. Optimizing chemical reactions with deep
  reinforcement learning. \emph{ACS central science} \textbf{2017}, \emph{3},
  1337--1344\relax
\mciteBstWouldAddEndPuncttrue
\mciteSetBstMidEndSepPunct{\mcitedefaultmidpunct}
{\mcitedefaultendpunct}{\mcitedefaultseppunct}\relax
\EndOfBibitem
\bibitem[Faber \latin{et~al.}(2017)Faber, Hutchison, Huang, Gilmer, Schoenholz,
  Dahl, Vinyals, Kearnes, Riley, and von Lilienfeld]{faber2017prediction}
Faber,~F.~A.; Hutchison,~L.; Huang,~B.; Gilmer,~J.; Schoenholz,~S.~S.
  \latin{et~al.}  Prediction errors of molecular machine learning models lower
  than hybrid DFT error. \emph{Journal of chemical theory and computation}
  \textbf{2017}, \emph{13}, 5255--5264\relax
\mciteBstWouldAddEndPuncttrue
\mciteSetBstMidEndSepPunct{\mcitedefaultmidpunct}
{\mcitedefaultendpunct}{\mcitedefaultseppunct}\relax
\EndOfBibitem
\bibitem[Gilmer \latin{et~al.}(2017)Gilmer, Schoenholz, Riley, Vinyals, and
  Dahl]{gilmer2017neural}
Gilmer,~J.; Schoenholz,~S.~S.; Riley,~P.~F.; Vinyals,~O.; Dahl,~G.~E. Neural
  message passing for quantum chemistry. \emph{arXiv preprint arXiv:1704.01212}
  \textbf{2017}, \relax
\mciteBstWouldAddEndPunctfalse
\mciteSetBstMidEndSepPunct{\mcitedefaultmidpunct}
{}{\mcitedefaultseppunct}\relax
\EndOfBibitem
\bibitem[Sch{\"u}tt \latin{et~al.}(2017)Sch{\"u}tt, Kindermans, Felix, Chmiela,
  Tkatchenko, and M{\"u}ller]{schutt2017schnet}
Sch{\"u}tt,~K.; Kindermans,~P.-J.; Felix,~H. E.~S.; Chmiela,~S.; Tkatchenko,~A.
  \latin{et~al.}  SchNet: A continuous-filter convolutional neural network for
  modeling quantum interactions. Advances in Neural Information Processing
  Systems. 2017; pp 991--1001\relax
\mciteBstWouldAddEndPuncttrue
\mciteSetBstMidEndSepPunct{\mcitedefaultmidpunct}
{\mcitedefaultendpunct}{\mcitedefaultseppunct}\relax
\EndOfBibitem
\bibitem[Sch{\"u}tt \latin{et~al.}(2017)Sch{\"u}tt, Arbabzadah, Chmiela,
  M{\"u}ller, and Tkatchenko]{schutt2017quantum}
Sch{\"u}tt,~K.~T.; Arbabzadah,~F.; Chmiela,~S.; M{\"u}ller,~K.~R.;
  Tkatchenko,~A. Quantum-chemical insights from deep tensor neural networks.
  \emph{Nature communications} \textbf{2017}, \emph{8}, 13890\relax
\mciteBstWouldAddEndPuncttrue
\mciteSetBstMidEndSepPunct{\mcitedefaultmidpunct}
{\mcitedefaultendpunct}{\mcitedefaultseppunct}\relax
\EndOfBibitem
\bibitem[Smith \latin{et~al.}(2017)Smith, Isayev, and Roitberg]{smith2017ani}
Smith,~J.~S.; Isayev,~O.; Roitberg,~A.~E. ANI-1: an extensible neural network
  potential with DFT accuracy at force field computational cost. \emph{Chemical
  science} \textbf{2017}, \emph{8}, 3192--3203\relax
\mciteBstWouldAddEndPuncttrue
\mciteSetBstMidEndSepPunct{\mcitedefaultmidpunct}
{\mcitedefaultendpunct}{\mcitedefaultseppunct}\relax
\EndOfBibitem
\bibitem[Feinberg \latin{et~al.}(2018)Feinberg, Sur, Husic, Mai, Li, Yang,
  Ramsundar, and Pande]{feinberg2018spatial}
Feinberg,~E.~N.; Sur,~D.; Husic,~B.~E.; Mai,~D.; Li,~Y. \latin{et~al.}  Spatial
  Graph Convolutions for Drug Discovery. \emph{arXiv preprint arXiv:1803.04465}
  \textbf{2018}, \relax
\mciteBstWouldAddEndPunctfalse
\mciteSetBstMidEndSepPunct{\mcitedefaultmidpunct}
{}{\mcitedefaultseppunct}\relax
\EndOfBibitem
\bibitem[Liu \latin{et~al.}(2017)Liu, Su, Han, Liu, Yang, Li, and
  Wang]{liu2017forging}
Liu,~Z.; Su,~M.; Han,~L.; Liu,~J.; Yang,~Q. \latin{et~al.}  Forging the basis
  for developing protein--ligand interaction scoring functions. \emph{Accounts
  of chemical research} \textbf{2017}, \emph{50}, 302--309\relax
\mciteBstWouldAddEndPuncttrue
\mciteSetBstMidEndSepPunct{\mcitedefaultmidpunct}
{\mcitedefaultendpunct}{\mcitedefaultseppunct}\relax
\EndOfBibitem
\bibitem[Hachmann \latin{et~al.}(2011)Hachmann, Olivares-Amaya, Atahan-Evrenk,
  Amador-Bedolla, S{\'a}nchez-Carrera, Gold-Parker, Vogt, Brockway, and
  Aspuru-Guzik]{hachmann2011harvard}
Hachmann,~J.; Olivares-Amaya,~R.; Atahan-Evrenk,~S.; Amador-Bedolla,~C.;
  S{\'a}nchez-Carrera,~R.~S. \latin{et~al.}  The Harvard clean energy project:
  large-scale computational screening and design of organic photovoltaics on
  the world community grid. \emph{The Journal of Physical Chemistry Letters}
  \textbf{2011}, \emph{2}, 2241--2251\relax
\mciteBstWouldAddEndPuncttrue
\mciteSetBstMidEndSepPunct{\mcitedefaultmidpunct}
{\mcitedefaultendpunct}{\mcitedefaultseppunct}\relax
\EndOfBibitem
\bibitem[Mysinger \latin{et~al.}(2012)Mysinger, Carchia, Irwin, and
  Shoichet]{mysinger2012directory}
Mysinger,~M.~M.; Carchia,~M.; Irwin,~J.~J.; Shoichet,~B.~K. Directory of useful
  decoys, enhanced (DUD-E): better ligands and decoys for better benchmarking.
  \emph{Journal of medicinal chemistry} \textbf{2012}, \emph{55},
  6582--6594\relax
\mciteBstWouldAddEndPuncttrue
\mciteSetBstMidEndSepPunct{\mcitedefaultmidpunct}
{\mcitedefaultendpunct}{\mcitedefaultseppunct}\relax
\EndOfBibitem
\bibitem[Gal(2016)]{gal2016uncertainty}
Gal,~Y. Uncertainty in deep learning. \emph{University of Cambridge}
  \textbf{2016}, \relax
\mciteBstWouldAddEndPunctfalse
\mciteSetBstMidEndSepPunct{\mcitedefaultmidpunct}
{}{\mcitedefaultseppunct}\relax
\EndOfBibitem
\bibitem[Begoli \latin{et~al.}(2019)Begoli, Bhattacharya, and
  Kusnezov]{begoli2019need}
Begoli,~E.; Bhattacharya,~T.; Kusnezov,~D. The need for uncertainty
  quantification in machine-assisted medical decision making. \emph{Nature
  Machine Intelligence} \textbf{2019}, \emph{1}, 20\relax
\mciteBstWouldAddEndPuncttrue
\mciteSetBstMidEndSepPunct{\mcitedefaultmidpunct}
{\mcitedefaultendpunct}{\mcitedefaultseppunct}\relax
\EndOfBibitem
\bibitem[McAllister \latin{et~al.}(2017)McAllister, Gal, Kendall, Van Der~Wilk,
  Shah, Cipolla, and Weller]{mcallister2017concrete}
McAllister,~R.; Gal,~Y.; Kendall,~A.; Van Der~Wilk,~M.; Shah,~A. \latin{et~al.}
   Concrete problems for autonomous vehicle safety: advantages of Bayesian deep
  learning. 2017\relax
\mciteBstWouldAddEndPuncttrue
\mciteSetBstMidEndSepPunct{\mcitedefaultmidpunct}
{\mcitedefaultendpunct}{\mcitedefaultseppunct}\relax
\EndOfBibitem
\bibitem[Kendall and Gal(2017)Kendall, and Gal]{kendall2017uncertainties}
Kendall,~A.; Gal,~Y. What uncertainties do we need in bayesian deep learning
  for computer vision? Advances in neural information processing systems. 2017;
  pp 5574--5584\relax
\mciteBstWouldAddEndPuncttrue
\mciteSetBstMidEndSepPunct{\mcitedefaultmidpunct}
{\mcitedefaultendpunct}{\mcitedefaultseppunct}\relax
\EndOfBibitem
\bibitem[Kwon \latin{et~al.}(2018)Kwon, Won, Kim, and
  Paik]{kwon2018uncertainty}
Kwon,~Y.; Won,~J.-H.; Kim,~B.~J.; Paik,~M.~C. Uncertainty quantification using
  Bayesian neural networks in classification: Application to ischemic stroke
  lesion segmentation. international conference on medical imaging with deep
  learning. 2018\relax
\mciteBstWouldAddEndPuncttrue
\mciteSetBstMidEndSepPunct{\mcitedefaultmidpunct}
{\mcitedefaultendpunct}{\mcitedefaultseppunct}\relax
\EndOfBibitem
\bibitem[Der~Kiureghian and Ditlevsen(2009)Der~Kiureghian, and
  Ditlevsen]{der2009aleatory}
Der~Kiureghian,~A.; Ditlevsen,~O. Aleatory or epistemic? Does it matter?
  \emph{Structural Safety} \textbf{2009}, \emph{31}, 105--112\relax
\mciteBstWouldAddEndPuncttrue
\mciteSetBstMidEndSepPunct{\mcitedefaultmidpunct}
{\mcitedefaultendpunct}{\mcitedefaultseppunct}\relax
\EndOfBibitem
\bibitem[Duvenaud \latin{et~al.}(2015)Duvenaud, Maclaurin, Iparraguirre,
  Bombarell, Hirzel, Aspuru-Guzik, and Adams]{duvenaud2015convolutional}
Duvenaud,~D.~K.; Maclaurin,~D.; Iparraguirre,~J.; Bombarell,~R.; Hirzel,~T.
  \latin{et~al.}  Convolutional networks on graphs for learning molecular
  fingerprints. Advances in neural information processing systems. 2015; pp
  2224--2232\relax
\mciteBstWouldAddEndPuncttrue
\mciteSetBstMidEndSepPunct{\mcitedefaultmidpunct}
{\mcitedefaultendpunct}{\mcitedefaultseppunct}\relax
\EndOfBibitem
\bibitem[Kipf and Welling(2016)Kipf, and Welling]{kipf2016semi}
Kipf,~T.~N.; Welling,~M. Semi-supervised classification with graph
  convolutional networks. \emph{arXiv preprint arXiv:1609.02907} \textbf{2016},
  \relax
\mciteBstWouldAddEndPunctfalse
\mciteSetBstMidEndSepPunct{\mcitedefaultmidpunct}
{}{\mcitedefaultseppunct}\relax
\EndOfBibitem
\bibitem[Ryu \latin{et~al.}(2018)Ryu, Lim, and Kim]{ryu2018deeply}
Ryu,~S.; Lim,~J.; Kim,~W.~Y. Deeply learning molecular structure-property
  relationships using graph attention neural network. \emph{arXiv preprint
  arXiv:1805.10988} \textbf{2018}, \relax
\mciteBstWouldAddEndPunctfalse
\mciteSetBstMidEndSepPunct{\mcitedefaultmidpunct}
{}{\mcitedefaultseppunct}\relax
\EndOfBibitem
\bibitem[Irwin and Shoichet(2005)Irwin, and Shoichet]{irwin2005zinc}
Irwin,~J.~J.; Shoichet,~B.~K. ZINC- A free database of commercially available
  compounds for virtual screening. \emph{Journal of chemical information and
  modeling} \textbf{2005}, \emph{45}, 177--182\relax
\mciteBstWouldAddEndPuncttrue
\mciteSetBstMidEndSepPunct{\mcitedefaultmidpunct}
{\mcitedefaultendpunct}{\mcitedefaultseppunct}\relax
\EndOfBibitem
\bibitem[Blundell \latin{et~al.}(2015)Blundell, Cornebise, Kavukcuoglu, and
  Wierstra]{blundell2015weight}
Blundell,~C.; Cornebise,~J.; Kavukcuoglu,~K.; Wierstra,~D. Weight uncertainty
  in neural networks. \emph{arXiv preprint arXiv:1505.05424} \textbf{2015},
  \relax
\mciteBstWouldAddEndPunctfalse
\mciteSetBstMidEndSepPunct{\mcitedefaultmidpunct}
{}{\mcitedefaultseppunct}\relax
\EndOfBibitem
\bibitem[Graves(2011)]{graves2011practical}
Graves,~A. Practical variational inference for neural networks. Advances in
  neural information processing systems. 2011; pp 2348--2356\relax
\mciteBstWouldAddEndPuncttrue
\mciteSetBstMidEndSepPunct{\mcitedefaultmidpunct}
{\mcitedefaultendpunct}{\mcitedefaultseppunct}\relax
\EndOfBibitem
\bibitem[Louizos and Welling(2017)Louizos, and
  Welling]{louizos2017multiplicative}
Louizos,~C.; Welling,~M. Multiplicative normalizing flows for variational
  bayesian neural networks. \emph{arXiv preprint arXiv:1703.01961}
  \textbf{2017}, \relax
\mciteBstWouldAddEndPunctfalse
\mciteSetBstMidEndSepPunct{\mcitedefaultmidpunct}
{}{\mcitedefaultseppunct}\relax
\EndOfBibitem
\bibitem[Srivastava \latin{et~al.}(2014)Srivastava, Hinton, Krizhevsky,
  Sutskever, and Salakhutdinov]{srivastava2014dropout}
Srivastava,~N.; Hinton,~G.; Krizhevsky,~A.; Sutskever,~I.; Salakhutdinov,~R.
  Dropout: a simple way to prevent neural networks from overfitting. \emph{The
  Journal of Machine Learning Research} \textbf{2014}, \emph{15},
  1929--1958\relax
\mciteBstWouldAddEndPuncttrue
\mciteSetBstMidEndSepPunct{\mcitedefaultmidpunct}
{\mcitedefaultendpunct}{\mcitedefaultseppunct}\relax
\EndOfBibitem
\bibitem[Gal and Ghahramani(2016)Gal, and Ghahramani]{gal2016dropout}
Gal,~Y.; Ghahramani,~Z. Dropout as a Bayesian approximation: Representing model
  uncertainty in deep learning. international conference on machine learning.
  2016; pp 1050--1059\relax
\mciteBstWouldAddEndPuncttrue
\mciteSetBstMidEndSepPunct{\mcitedefaultmidpunct}
{\mcitedefaultendpunct}{\mcitedefaultseppunct}\relax
\EndOfBibitem
\bibitem[Battaglia \latin{et~al.}(2018)Battaglia, Hamrick, Bapst,
  Sanchez-Gonzalez, Zambaldi, Malinowski, Tacchetti, Raposo, Santoro, Faulkner,
  \latin{et~al.} others]{battaglia2018relational}
Battaglia,~P.~W.; Hamrick,~J.~B.; Bapst,~V.; Sanchez-Gonzalez,~A.; Zambaldi,~V.
  \latin{et~al.}  Relational inductive biases, deep learning, and graph
  networks. \emph{arXiv preprint arXiv:1806.01261} \textbf{2018}, \relax
\mciteBstWouldAddEndPunctfalse
\mciteSetBstMidEndSepPunct{\mcitedefaultmidpunct}
{}{\mcitedefaultseppunct}\relax
\EndOfBibitem
\bibitem[Vaswani \latin{et~al.}(2017)Vaswani, Shazeer, Parmar, Uszkoreit,
  Jones, Gomez, Kaiser, and Polosukhin]{vaswani2017attention}
Vaswani,~A.; Shazeer,~N.; Parmar,~N.; Uszkoreit,~J.; Jones,~L. \latin{et~al.}
  Attention is all you need. Advances in Neural Information Processing Systems.
  2017; pp 5998--6008\relax
\mciteBstWouldAddEndPuncttrue
\mciteSetBstMidEndSepPunct{\mcitedefaultmidpunct}
{\mcitedefaultendpunct}{\mcitedefaultseppunct}\relax
\EndOfBibitem
\bibitem[Velickovic \latin{et~al.}(2017)Velickovic, Cucurull, Casanova, Romero,
  Lio, and Bengio]{velickovic2017graph}
Velickovic,~P.; Cucurull,~G.; Casanova,~A.; Romero,~A.; Lio,~P. \latin{et~al.}
  Graph attention networks. \emph{arXiv preprint arXiv:1710.10903}
  \textbf{2017}, \relax
\mciteBstWouldAddEndPunctfalse
\mciteSetBstMidEndSepPunct{\mcitedefaultmidpunct}
{}{\mcitedefaultseppunct}\relax
\EndOfBibitem
\bibitem[Gal \latin{et~al.}(2017)Gal, Hron, and Kendall]{gal2017concrete}
Gal,~Y.; Hron,~J.; Kendall,~A. Concrete dropout. Advances in Neural Information
  Processing Systems. 2017; pp 3581--3590\relax
\mciteBstWouldAddEndPuncttrue
\mciteSetBstMidEndSepPunct{\mcitedefaultmidpunct}
{\mcitedefaultendpunct}{\mcitedefaultseppunct}\relax
\EndOfBibitem
\bibitem[Kingma and Ba(2014)Kingma, and Ba]{kingma2014adam}
Kingma,~D.~P.; Ba,~J. Adam: A method for stochastic optimization. \emph{arXiv
  preprint arXiv:1412.6980} \textbf{2014}, \relax
\mciteBstWouldAddEndPunctfalse
\mciteSetBstMidEndSepPunct{\mcitedefaultmidpunct}
{}{\mcitedefaultseppunct}\relax
\EndOfBibitem
\bibitem[Landrum(2006)]{landrum2006rdkit}
Landrum,~G. RDKit: Open-source cheminformatics. 2006\relax
\mciteBstWouldAddEndPuncttrue
\mciteSetBstMidEndSepPunct{\mcitedefaultmidpunct}
{\mcitedefaultendpunct}{\mcitedefaultseppunct}\relax
\EndOfBibitem
\bibitem[Scharber \latin{et~al.}(2006)Scharber, M{\"u}hlbacher, Koppe, Denk,
  Waldauf, Heeger, and Brabec]{scharber2006design}
Scharber,~M.~C.; M{\"u}hlbacher,~D.; Koppe,~M.; Denk,~P.; Waldauf,~C.
  \latin{et~al.}  Design rules for donors in bulk-heterojunction solar
  cells—Towards 10\% energy-conversion efficiency. \emph{Advanced materials}
  \textbf{2006}, \emph{18}, 789--794\relax
\mciteBstWouldAddEndPuncttrue
\mciteSetBstMidEndSepPunct{\mcitedefaultmidpunct}
{\mcitedefaultendpunct}{\mcitedefaultseppunct}\relax
\EndOfBibitem
\bibitem[Lopez \latin{et~al.}(2016)Lopez, Pyzer-Knapp, Simm, Lutzow, Li,
  Seress, Hachmann, and Aspuru-Guzik]{lopez2016harvard}
Lopez,~S.~A.; Pyzer-Knapp,~E.~O.; Simm,~G.~N.; Lutzow,~T.; Li,~K.
  \latin{et~al.}  The Harvard organic photovoltaic dataset. \emph{Scientific
  data} \textbf{2016}, \emph{3}, 160086\relax
\mciteBstWouldAddEndPuncttrue
\mciteSetBstMidEndSepPunct{\mcitedefaultmidpunct}
{\mcitedefaultendpunct}{\mcitedefaultseppunct}\relax
\EndOfBibitem
\bibitem[Zhao and Truhlar(2008)Zhao, and Truhlar]{zhao2008m06}
Zhao,~Y.; Truhlar,~D.~G. The M06 suite of density functionals for main group
  thermochemistry, thermochemical kinetics, noncovalent interactions, excited
  states, and transition elements: two new functionals and systematic testing
  of four M06-class functionals and 12 other functionals. \emph{Theoretical
  Chemistry Accounts} \textbf{2008}, \emph{120}, 215--241\relax
\mciteBstWouldAddEndPuncttrue
\mciteSetBstMidEndSepPunct{\mcitedefaultmidpunct}
{\mcitedefaultendpunct}{\mcitedefaultseppunct}\relax
\EndOfBibitem
\end{mcitethebibliography}
\end{document}